# Disambiguation of Super Parts of Speech (or Supertags): Almost Parsing

Aravind K. Joshi and B. Srinivas

Department of Computer and Information Science
University of Pennsylvania
Philadelphia, PA 19104, USA

{joshi, srini}@linc.cis.upenn.edu

**Abstract:** In a lexicalized grammar formalism such as Lexicalized Tree-Adjoining Grammar (LTAG), each lexical item is associated with at least one elementary structure (supertag) that localizes syntactic and semantic dependencies. Thus a parser for a lexicalized grammar must search a large set of supertags to choose the right ones to combine for the parse of the sentence. We present techniques for disambiguating supertags using local information such as lexical preference and local lexical dependencies. The similarity between LTAG and Dependency grammars is exploited in the dependency model of supertag disambiguation. The performance results for various models of supertag disambiguation such as unigram, trigram and dependency-based models are presented.

## 1 Introduction

Part-of-speech disambiguation techniques (**taggers**) are often used to eliminate (or substantially reduce) the part-of-speech ambiguity prior to parsing. The taggers are all local in the sense that they use information from a limited context in deciding which tag(s) to choose for each word. As is well known, these taggers are quite successful.

In a lexicalized grammar such as the Lexicalized Tree-Adjoining Grammar (LTAG), each lexical item is associated with at least one elementary structure (tree). The elementary structures of LTAG localize dependencies, including long distance dependencies, by requiring that all and only the dependent elements be present within the same structure. As a result of this localization, a lexical item may be (and, in general, almost always is) associated with more than one elementary structure. We will call these elementary structures **supertags**, in order to distinguish them from the standard part-of-speech tags. Note that even when a word has a unique standard part-of-speech, say a verb (V), there will usually be more than one supertag associated with this word. Since when the parse is complete, there is only one supertag for each word (assuming there is no global ambiguity), an LTAG parser (Schabes, 1988) needs to search a large space of supertags to select the right one for each word before combining them for the parse of a sentence. It is this problem of supertag disambiguation that we address in this paper.

Since LTAGs are lexicalized, we are presented with a novel opportunity to eliminate or substantially reduce the supertag assignment ambiguity by using local information such as local lexical dependencies, prior to parsing. As in standard part-of-speech disambiguation, we can use local statistical information in the form of n-gram models based on the distribution of supertags in a LTAG parsed corpus. Moreover, since the supertags encode dependency information, we can also use information about the distribution of distances between a given supertag and its dependent supertags.

Note that as in standard part-of-speech disambiguation, supertag disambiguation could have been done by a parser. However, carrying out part-of-speech disambiguation prior to parsing makes the job of the parser much easier and therefore speeds it up. Supertag disambiguation as proposed in this paper reduces the work of the parser even further. After supertag disambiguation, we have effectively completed the parse and the parser need 'only' combine the individual structures; hence the term–*almost parsing*. This method can also be used to parse sentence fragments in cases where the supertag sequence after the disambiguation may not combine into a single structure.

The main goal of this paper is to present techniques for disambiguating supertags, and to evaluate their performance and their impact on LTAG parsing. Although presented with respect to LTAG, these techniques are applicable to lexicalized grammars in general. Section 2 provides an introduction to Lexicalized Tree Adjoining Grammars. The objective of supertag disambiguation is illustrated through an example in Section 3. Section 4 briefly describes the system used to collect the data needed for supertag disambiguation. Various methods and their performance results for supertag disambiguation are discussed in detail in Section 5.

## 2 Lexicalized Tree Adjoining Grammars

Lexicalized Tree Adjoining Grammar (LTAG) is a lexicalized tree rewriting grammar formalism. The primary structures of LTAG are ELEMENTARY TREES. Each elementary tree has a lexical item (anchor) on its frontier and provides an extended domain of locality over which the anchor specifies syntactic and semantic (predicate-argument) constraints. Elementary trees are of two kinds: INITIAL TREES and AUXILIARY TREES. Examples of initial trees ($\alpha$s) and auxiliary trees ($\beta$s) are shown in Figure 1. Nodes on the frontier of initial trees are marked as substitution sites by a '↓', while exactly one node on the frontier of an auxiliary tree, whose label matches the label of the root of the tree, is marked as a foot node by a '*'. The other nodes on the frontier of an auxiliary tree are marked as substitution sites. LTAG factors recursion from the statement of the syntactic dependencies. Elementary trees (initial and auxiliary) are the domain for specifying dependencies. Recursion is specified via the auxiliary trees. Elementary trees are combined by the **Substitution** and **Adjunction** operations. Substitution inserts elementary trees at the substitution nodes of other elementary trees. Adjunction inserts auxiliary trees into elementary trees at the node whose label is the same as the root label of the auxiliary tree. As an example, the component trees ($\alpha_8, \alpha_2, \alpha_3, \alpha_4, \beta_8, \alpha_5, \alpha_6$), shown in Figure 1 can be combined to form the sentence *John saw a man with the telescope*[1] as follows:

1. $\alpha_8$ substitutes at the $NP_0$ node in $\alpha_2$.

2. $\alpha_3$ substitutes at the DetP node in $\alpha_4$, the result of which is substituted at the $NP_1$ node in $\alpha_2$.

3. $\alpha_5$ substitutes at the DetP node in $\alpha_6$, the result of which is substituted at the NP node in $\beta_8$.

4. The result of step (3) above adjoins to the VP node of the result of step (2). The resulting parse tree is shown in Figure 2(a).

The process of combining the elementary trees resulting in the parse of the sentence is represented by the **derivation tree**, shown in Figure 2(b). The nodes of the derivation tree are the tree names that are anchored by the appropriate lexical item. The composition operation is indicated by the nature of the arcs – dashed line for substitution and bold line for adjunction, while the address of the operation is indicated as part of the node label. The derivation tree can also be interpreted as a dependency graph with unlabeled arcs between words of the sentence as shown in Figure 2(c).

We will call the elementary structures associated with each lexical item as super parts-of-speech (super POS) or **supertags**.

## 3 Example of Supertagging

As a result of localization in LTAG, a lexical item may be associated with more than one supertag. The example in Figure 3 illustrates the initial set of supertags assigned to each word of the sentence *John saw a man with the telescope*. The order of the supertags for each lexical item in the example is not significant. Figure 3 also shows the final supertag sequence assigned by the supertagger, which picks the best supertag sequence using statistical information (described in Section 4) about individual supertags and their dependencies on other supertags. The chosen supertags are combined to derive a parse, as explained in Section 2.

---
[1]The parse with the PP attached to the NP has not been shown.

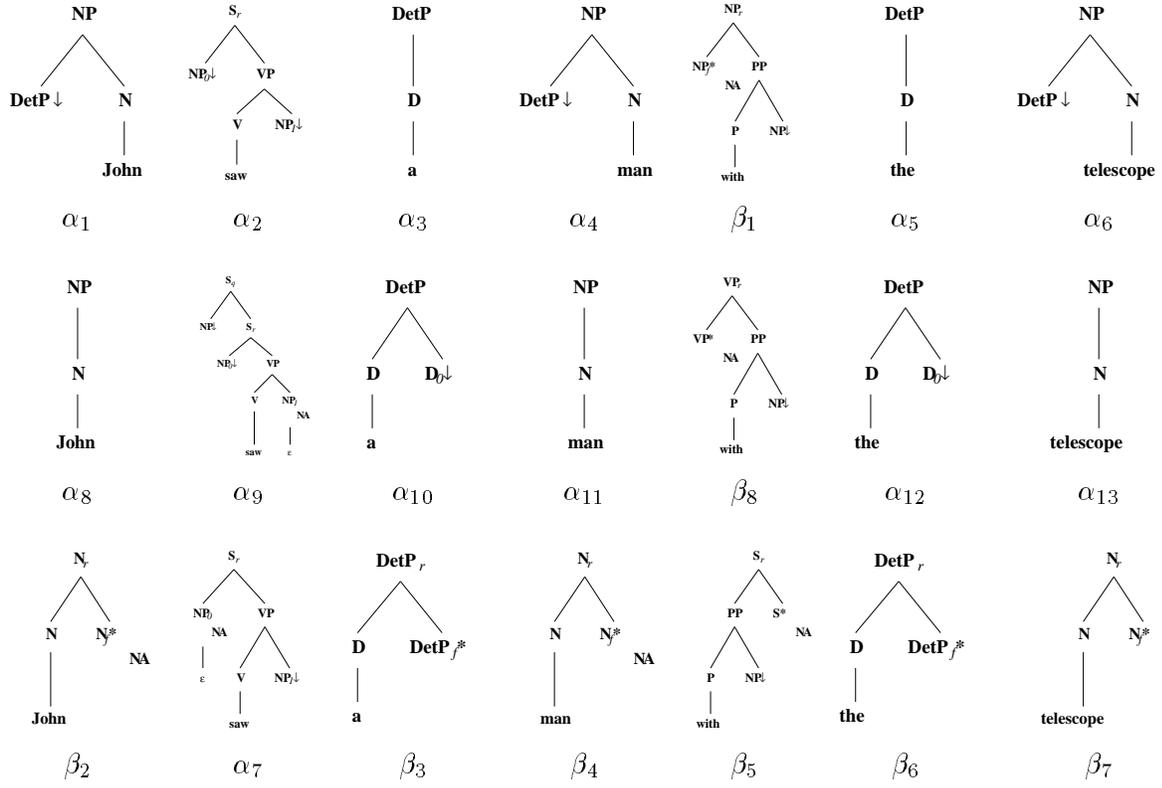

Figure 1: **Elementary trees of LTAG**

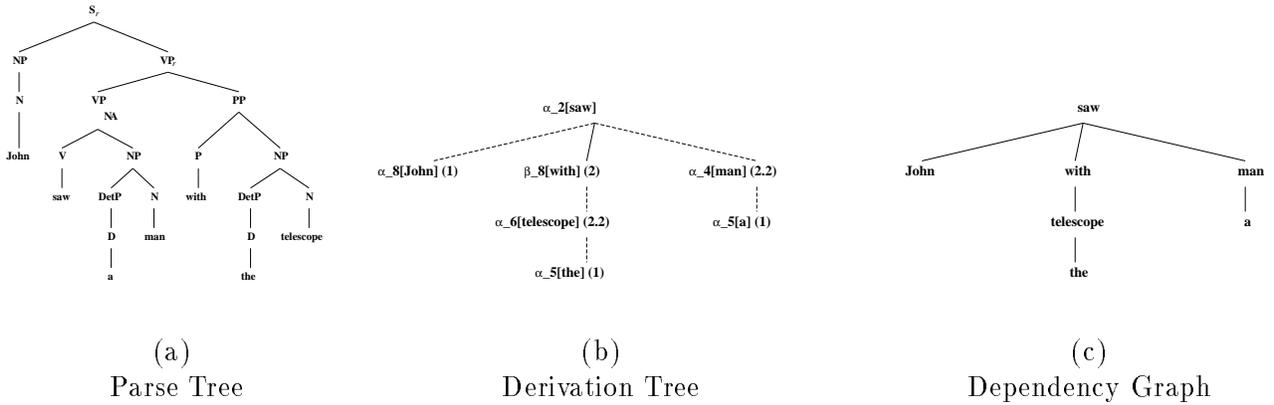

Figure 2: **Structures of LTAG**

| Sentence: | John | saw | a | man | with | the | telescope. |
|---|---|---|---|---|---|---|---|
| Initial Supertag set: | $\alpha_1$ | $\alpha_2$ | $\alpha_3$ | $\alpha_4$ | $\beta_1$ | $\alpha_5$ | $\alpha_6$ |
| | $\alpha_8$ | $\alpha_9$ | $\alpha_{10}$ | $\alpha_{11}$ | $\beta_8$ | $\alpha_{12}$ | $\alpha_{13}$ |
| | $\beta_2$ | $\alpha_7$ | $\beta_3$ | $\beta_4$ | $\beta_5$ | $\beta_6$ | $\beta_7$ |
| Final Assignment: | $\alpha_8$ | $\alpha_2$ | $\alpha_3$ | $\alpha_4$ | $\beta_8$ | $\alpha_5$ | $\alpha_6$ |

Figure 3: **Supertag Assignment for** *John saw a man with the telescope*

Without the supertagger, the parser would have to process combinations of the entire set of trees (28); with it the parser must only processes combinations of 7 trees.

## 4 Data Collection

The data required for disambiguating supertags (discussed in Section 5) have been collected by parsing the Wall Street Journal[2], IBM-manual and ATIS corpora using the wide-coverage English grammar being developed as part of the XTAG system (Doran et. al., 1994). The parses generated by the system for these sentences from the corpora are not subjected to any kind of filtering or selection. All the derivation structures are used in the collection of the statistics.

### 4.1 About XTAG

XTAG is a large ongoing project to develop a wide-coverage grammar for English, based on the LTAG formalism. It also serves as an LTAG grammar development system and consists of a predictive left-to-right parser, an X-window interface, a morphological analyzer and a part-of-speech tagger. The wide-coverage English grammar of the XTAG system contains 317,000 inflected items in the morphology (213,000 of these are nouns and 46,500 are verbs) and 37,000 entries in the syntactic lexicon. The syntactic lexicon associates words with the trees that they anchor. There are 385 trees in all, in a grammar which is composed of 40 different subcategorization frames. Each word in the syntactic lexicon, on the average, depending on the standard parts-of-speech of the word, is an anchor for about 8 to 40 elementary trees.

## 5 Models, Experiments and Results

The supertag statistics which have been used in the preliminary experiments described below have been collected from the XTAG parsed corpora. The derivation structures resulting from parsed corpora (Wall Street Journal, for the experiments described here) serve as training data for these experiments.

[2]Sentences of length $\leq 15$ words

### 5.1 Unigram model

One method of disambiguating the supertags assigned to each word is to order the supertags by the lexical preference that the word has for them. The frequency with which a certain supertag is associated with a word is a direct measure of its lexical preference for that supertag. Associating frequencies with the supertags and using them to associate a particular supertag with a word is clearly the simplest means of disambiguating supertags. Thus,
Supertag$(w_i) = t_k \ni \text{argmax}_{t_k} \text{unigram}(t_k \mid w_i)$.

#### 5.1.1 Experiments and Results

Owing to sparseness of data, we have backed-off from word/supertag pairs to part-of-speech/supertag pairs, i.e., collected the unigram frequencies of supertags associated with the part-of-speech assigned to words instead of the words themselves. Table 1 illustrates the nature of the statistics used, with a few sample entries.

| Part-of-speech | (supertag, unigram probability) |
|---|---|
| N | ($\alpha_1$, 0.218) <br> ($\alpha_8$, 0.375) <br> ($\beta_2$, 0.282) |
| V | ($\alpha_2$, 0.099) |
| D | ($\alpha_3$, 0.963) |

Table 1: Sample entries of unigram database

| Top $n$ Supertags | % Success |
|---|---|
| $n = 1$ | 15% |
| $n = 2$ | 22% |
| $n = 3$ | 52% |

Table 2: Results from the Unigram Supertag Model

The words are first assigned standard parts-of-speech using a conventional tagger (Church, 1988). Then the set of supertags associated with each word is retrieved from XTAG's syntactic database. These supertags are ordered based on their unigram frequency, and the top $n$ supertags are associated with the word. Table 2 summarizes the success percentage on a held out test set of 100 Wall Street Journal sentences, as $n$ is varied. If a sentence parses using the $n$ supertags selected for each word then the assignment is considered a success.

The unigram supertagger that selects top three supertags has been interfaced with XTAG. This speeds the runtime of the parser by 87% on the average, whenever the supertagger succeeds.

## 5.2 n-gram model

In a unigram model a word is always associated with the supertag that is most preferred by the word, irrespective of the context in which the word appears. An alternate method that is sensitive to context is the n-gram model. The n-gram model takes into account the contextual dependency probabilities between supertags within a window of $n$ words in associating supertags with words. Thus the most probable supertag sequence for a $N$ word sentence is given by

$\hat{T}$ = argmax$_T$ Pr($T_1,T_2,\ldots,T_N$) *
$\quad\quad\quad\quad$ Pr($W_1,W_2,\ldots,W_N|T_1,T_2,\ldots,T_N$)

To compute this using only local information, we approximate, taking the probability of a word to depend only on its supertag

Pr($W_1,W_2,\ldots,W_N|T_1,T_2,\ldots,T_N$)
$$\approx \prod_{i=1}^{N} \text{Pr}(W_i \mid T_i)$$

and also use an n-gram (trigram, in this case) approximation

$\quad$ Pr($T_1,T_2,\ldots,T_N$) $\approx \prod_{i=1}^{N}$ Pr($T_i \mid T_{i-2}, T_{i-1}$)

### 5.2.1 Experiments and Results

A trigram model has been used to model the contextual dependencies in supertag sequences. Again, due to sparseness of data, the particular words have been ignored and the training of the trigram model has been done on the part-of-speech/supertag pair. The model has been tested on the same set of held out sentences as in the unigram experiment. The percentage success is 68%, i.e., 68% of the words of the test corpus were assigned the correct supertag.

## 5.3 Dependency model

In the n-gram model for disambiguating supertags, dependencies between supertags that appear beyond the $n$ word window cannot be incorporated into the model. This limitation can be overcome if no a priori bound is set on the size of the window but instead a probability distribution of the distances of the dependent supertags for each supertag is maintained. A supertag is dependent on another supertag if the former substitutes or adjoins into the latter[3].

### 5.3.1 Experiments and Results

Table 3 shows the data required for the dependency model of supertag disambiguation. Ideally each entry would be indexed by a (word, supertag) pair but, due to sparseness of data, we have backed-off to a (POS, supertag) pair. Each entry contains the following information.

- POS and Supertag pair.

- List of + and −, representing the direction of the dependent supertags with respect to the indexed supertag. (Size of this list indicates the total number of dependent supertags required.)

- Dependent supertag.

- Signed number representing the direction and the ordinal position of the particular dependent supertag mentioned in the entry from the position of the indexed supertag.

- A probability of occurrence of such a dependency. The sum probability over all the dependent supertags at all ordinal positions in the same direction is one.

For example, the fourth entry in the Table 3 reads that the tree $\alpha_2$, anchored by a verb (V), has a left and a right dependent (−, +) and the first word to the left (−1), with the tree $\alpha_8$, is dependent on the current word. The strength of this association is represented by the probability 0.300.

The dependency model of disambiguation works as follows. Suppose $\alpha_2$ is a member of the set of supertags associated with a word at position $n$ in the sentence. The algorithm proceeds to satisfy the dependency requirement of $\alpha_2$ by picking up the dependency entries for each of the directions. It picks a dependency data entry (the fourth entry, say) from the database that is indexed by $\alpha_2$ and proceeds to set up a path with the first word to the left that has the dependent supertag ($\alpha_8$) as a member of its set of supertags. If the first word to the left that has $\alpha_8$ as a member of its set of supertags is at position $m$, then an arc is set up between $\alpha_2$ and $\alpha_8$. Also, the arc is verified not to kite-string-tangle[4] with any other arcs in the path up to $\alpha_2$. The path probability up to $\alpha_2$ is incremented by $\log 0.300$ to reflect the success of the match. The path probability up to $\alpha_8$ incorporates the unigram probability of $\alpha_8$. On the other hand, if no word is found that has $\alpha_8$ as a member of its set of supertags then the entry is ignored. The algorithm makes a greedy choice by selecting the path with the maximum path probability to extend to the remaining directions in the dependency list. A successful supertag sequence is one which assigns a supertag to each position such that each supertag has all of its dependents and maximizes the accumulated path probability. It is to be noted that the algorithm when pairing the head and its dependent is not really parsing since it does so even without looking at the structure of the string between the head and the dependent.

---

[3] We are computing dependencies between words with respect to supertags associated with the words, although the complete structure of the supertags is not used. It is of interest to compare our work with some other dependency-based approaches as described by, for example, Sleator (Sleator and Temperley, 1990), Hindle (Hindle, 1993), Milward (Milward, 1992).

| (P.O.S,Supertag) | Direction of Dependent Supertag | Dependent Supertag | Ordinal position | Prob |
|---|---|---|---|---|
| (D,$\alpha_5$) | () | - | - | - |
| (N,$\alpha_8$) | () | - | - | - |
| (N,$\alpha_1$) | (−) | $\alpha_3$ | −1 | 0.999 |
| (V,$\alpha_2$) | (−, +) | $\alpha_8$ | −1 | 0.300 |
| (V,$\alpha_2$) | (−, +) | $\alpha_8$ | 1 | 0.374 |

Table 3: Dependency Data

The implementation and testing of this model of supertag disambiguation is underway. Table 4 shows preliminary results on the same held out test set of 100 Wall Street Journal sentences that was used in the unigram and trigram models. The table shows two measures of evaluation. In the first, the dependency link measure, the test sentences were independently hand tagged with dependency links and then were used to match the links output by the dependency model. The columns show the total number of dependency links in the hand tagged set, the number of matched links output by this model and the percentage correctness. The second measure, supertags, shows the total number of correct supertags assigned to the words in the corpus by this model.

| Criterion | Total number | Number correct | % correct |
|---|---|---|---|
| Dependency links | 815 | 620 | 76.07% |
| Supertags | 915 | 707 | 77.26% |

Table 4: Results of Dependency model

## 6 Conclusion

Lexicalized grammars associate with each word richer structures (trees in case of LTAGs and categories in case of Combinatory Categorial Grammars (CCGs)) over which the word specifies syntactic and semantic constraints. Hence every word is associated with a much larger set of more complex structures than in the case where the words are associated with standard parts-of-speech. However, these more complex descriptions allow more complex constraints to be imposed and verified locally on the contexts in which these words appear. This feature of lexicalized grammars can be taken advantage of, to further reduce the disambiguation task of the parser, as shown in supertag disambiguation. Hence supertag disambiguation can be used as a general pre-parsing component of lexicalized grammar parsers.

The degree of distinction between supertag disambiguation and parsing varies, depending on the lexicalized grammar being considered. For both LTAG and CCG, supertag disambiguation serves as a pre-parser filter that effectively weeds out inappropriate elementary structures (trees or categories) given the context of the sentence. It also indicates the dependencies among the elementary structures but not the specific operation to be used to combine the structures or the address at which the operation is to be performed – "an almost parse". In cases where the supertag sequence for the given input string cannot be combined to form a complete structure, the "almost parse" may indeed be the best one can do.

---
[4] Two arcs $(a,c)$ and $(b,d)$ kite-string-tangle if $a < b < c < d$ or $b < a < d < c$.

In case of LTAG, even though no explicit substitutions or adjunctions are shown, the dependencies among LTAG trees uniquely identify the combining operation between the trees and the node at which the operation can be performed is almost always unique[5]. Thus supertag disambiguation is almost parsing for LTAGs. In contrast, the dependencies among the CCG categories do not result in directly identifying the combining operations between the categories since two categories can often be combined in more than one way. Hence for CCG further processing needs to be performed to obtain the complete parse of the sentence, although without any supertag ambiguities.

The supertag disambiguation, dependency model in particular, is even closer to parsing in dependency grammar formalism. Dependency parsers establish relationships among words, unlike the phrase-structure parsers which construct a phrase-structure tree spanning the words of the input. Since LTAGs are lexicalized and each elementary tree is associated with at least one lexical item, the supertag disambiguation for LTAG can therefore be viewed as establishing the relationship[6] among words as dependency parsers do. Then the elementary structures that the related words anchor are combined to reconstruct the phrase-structure tree similar to the result of phrase-structure parsers. Thus the interplay of both dependency and phrase-structure grammars can be seen in LTAGs. Rambow and Joshi (Rambow and Joshi, 1993) discuss in greater detail the use of LTAG in relating dependency analyses to phrase-structure analyses and propose a dependency-based parser for a phrase-structure based grammar.

In summary, we have presented a new technique that performs the disambiguation of supertags using local information such as lexical preference and local lexical dependencies. This technique, like part-of-speech disambiguation, reduces the disambiguation task that needs to be done by the parser. After the disambiguation, we have effectively completed the parse of the sentence and the parser needs 'only' to complete the adjunction and substitutions. This method can also serve to parse sentence fragments in cases where the supertag sequence after the disambiguation may not combine to form a single structure. We have implemented this technique of disambiguation using the n-gram models using the probability data collected from LTAG parsed corpus. The similarity between LTAG and Dependency grammars is exploited in the dependency model of supertag disambiguation. The performance results of these models have been presented.

# References

Church, K. (1988). A Stochastic Parts Program and Noun Phrase Parser for Unrestricted Text. In *2nd Applied Natural Language Processing Conference 1988.*

Doran, C., Egedi, D., Hockey, B.A. and Srinivas, B. (1994). *XTAG Technical Report.* Department of Computer and Information Sciences, University of Pennsylvania, Philadelphia, PA. In progress

Hindle, D. (1993). Prediction of Lexicalized Tree Fragments in Text ARPA Workshop on Human Language Technology, March 1993.

Milward, D. (1992). Dynamics, Dependency Grammars and Incremental Interpretation. In *Proceedings of the 14th International Conference on Computational Linguistics (COLING'92)*, Nantes, France, August.

Rambow, O. and Joshi, A.K. (1993). Dependency Parsing for Phrase-Structure Grammars. *Manuscript*, University of Pennsylvania.

Sleator, D. and Temperley, D. (1991). Parsing English with a Link Grammar. *Technical report CMU-CS-91-196*, Department of Computer Science, Carnegie Mellon University, 1991.

Schabes, Y., Abeillé A. and Joshi, A.K. (1988). Parsing strategies with 'lexicalized' grammars: Application to tree adjoining grammars. In *Proceedings of the 12th International Conference on Computational Linguistics (COLING'88)*, Budapest, Hungary, August.

---

[5] In some cases, the dependency information between an auxiliary and an elementary tree may be insufficient to uniquely identify the address of adjunction, if the auxiliary tree can adjoin to more than one node in the elementary tree, since the specific attachments are not shown.

[6] The relational labels between two words in LTAG is associated with the address of the operation between the trees that the words anchor.

Schabes, Y. (1990). *Mathematical and Computational Aspects of Lexicalized Grammars*. Ph.D. thesis, University of Pennsylvania, Philadelphia, PA, August. Available as technical report (MS-CIS-90-48, LINC LAB179) from the Department of Computer and Information Science.